\documentclass[
]{ceurart}

\usepackage{amsmath, amssymb, amsfonts, bm, mathtools}
\usepackage{adjustbox, tabularx}
\usepackage{makecell, booktabs}
\usepackage{array, multirow}

\usepackage{graphicx}
\graphicspath{{./Metadata/}}
\usepackage{subcaption}
\usepackage{svg}
\usepackage{todonotes}
\usepackage{hyperref}
\usepackage{textcomp}
\usepackage{xcolor}

\usepackage[margin=3cm]{geometry}
\usepackage{algorithm2e}

\newcommand{\estimates}{\overset{\scriptscriptstyle\wedge}{=}}
\begin{document}

\copyrightyear{2023}
\copyrightclause{Copyright for this paper by its authors.
  Use permitted under Creative Commons License Attribution 4.0
  International (CC BY 4.0).}

\conference{EWAF'23: European Workshop on Algorithmic Fairness,
  June 07--09, 2023, Winterthur, Switzerland}

\title{Affinity Clustering Framework for Data Debiasing Using Pairwise Distribution Discrepancy}

\author[1,2]{Siamak Ghodsi}[%
orcid=0000-0002-3306-4233,
email=ghodsi@l3s.de,
url=https://siamakghodsi.github.io/,
]
\address[1]{L3S Research Center, Leibniz Universit\"at Hannover, Germany}
\address[2]{Freie Universit\"at Berlin, Dept. of Mathematics and Computer Science, Berlin, Germany}

\author[3]{Eirini Ntoutsi}[%
orcid=0000-0001-5729-1003,
email=eirini.ntoutsi@unibw.de,
]
\address[3]{Research Institute CODE, Bundeswehr University Munich, Germany}

\begin{abstract}
Group imbalance usually caused by insufficient or unrepresentative data collection procedures, is among the main reasons for the emergence of representation bias in datasets. Representation bias can exist with respect to different groups of one or more protected attributes and might lead to prejudicial and discriminatory outcomes toward certain groups of individuals; in case if a learning model is trained on such biased data. In this paper, we propose \emph{MASC} a data augmentation approach based on affinity clustering of existing data in similar datasets. An arbitrary target dataset utilizes protected group instances of other neighboring datasets that locate in the same cluster, in order to balance out the cardinality of its non-protected and protected groups. To form clusters where datasets can share instances for protected-group augmentation, an affinity clustering pipeline is developed based on an affinity matrix. The formation of the affinity matrix relies on computing the discrepancy of distributions between each pair of datasets and translating these discrepancies into a symmetric pairwise similarity matrix. Furthermore, a non-parametric spectral clustering is applied to the affinity matrix and the corresponding datasets are categorized into an optimal number of clusters automatically. 
We perform a step-by-step experiment as a demo of our method to both show the procedure of the proposed data augmentation method and also to evaluate and discuss its performance. In addition, a comparison to other data augmentation methods before and after the augmentations are provided as well as model evaluation performance analysis of each of the competitors compared to our method. In our experiments, bias is measured in a non-binary protected attribute setup w.r.t. \textit{racial} groups distribution for two separate minority groups in comparison with the majority group before and after debiasing. Empirical results imply that our method of augmenting dataset biases using real (genuine) data from similar contexts can effectively debias the target datasets comparably to existing data augmentation strategies.
  
\end{abstract}


\maketitle

\section{Introduction}
\label{sec:Intro}

Recent years have brought extraordinary advances in the field of Artificial Intelligence (AI) such that now AI-based technologies replace humans at many critical decision points, such as who will get a loan~\cite{DBLP:conf/icse/VermaR18} and who will get hired for a job~\cite{DBLP:journals/widm/NtoutsiFGINVRTP20}.
There are clear benefits to algorithmic decision-making; unlike people, machines do not become tired or bored \cite{DBLP:journals/csur/MehrabiMSLG21}, and can take into account orders of magnitude with more factors than people can. However, like people, data-driven algorithms are vulnerable to biases that render their decisions “unfair”. In automated decision-making, fairness is the absence of any prejudice or favoritism toward an individual or a group based on their inherent or acquired protected attributes such as `race' or `gender'. Thus, an unfair algorithm is one whose decisions are skewed toward a particular group of people.

One of the leading causes of unfair automated decisions in many real-world scenarios is due to unrepresentative, insufficient, or biased data fed to the learning algorithm \cite{DBLP:journals/csur/PessachS23}. Consequently, such biases can lead to certain discriminatory and prejudicial decisions harming sensitive groups e.g. racial/gender minorities in practice.
To overcome this issue, in this paper, we propose a mechanism for \textbf{M}inority \textbf{A}ugmentation of biased datasets coming from separate but similar sources of data (described in the same feature space) through a \textbf{S}pectral \textbf{C}lustering scheme (\textbf{MASC}). The method proposes a way to augment underrepresented minority groups of an arbitrary task (hereafter we use the terms dataset and task interchangeably) by increasing their instances from a subset of contextually similar datasets that belong to the same cluster. Our proposed method performs an affinity clustering based on distribution discrepancy (that is used as a distance measure) among tasks to group similar tasks into a pre-defined number of clusters. Within each cluster, any member dataset can use instances shared by neighboring (mutually most similar) tasks to augment their underrepresented groups as compensation for group cardinality difference (that leads to representation and imbalance bias) according to a protected attribute. 

Our main contributions can be summarized as follows:
\begin{itemize}
    \item A new data augmentation framework for data  debiasing towards statistical balancing between non-protected and protected group(s) based on most similar neighbors.
    \item Utilizing distribution shift metrics to quantify pairwise discrepancy between different datasets/ joint distributions 
    \item A spectral clustering framework to group similar datasets based on the discrepancy between the joint distribution of these datastes.
    \item Clustering into an optimal number of clusters using a graph theoretic heuristic, known as ''Eigen-gap or Spectral-gap`` to avoid parameter selection and thus avoid any additional bias in the pipeline.
\end{itemize}

The rest of the paper is organized as follows: 
Preliminaries, related works, and motivation are presented in Section~\ref{sec:related_works}. In Section \ref{sec:proposed}, we present the proposed data augmentation pipeline. Experimental evaluation results are provided in Section~\ref{sec:EXP}, including an intuitive example of applying the proposed method. Finally, Section~\ref{sec:conclusion} concludes this work, discusses its limitations and points out to future directions.

\section{Preliminaries and Related Works}
\label{sec:related_works}
In this section, the necessary theoretical background and a brief literature review of these necessary notions are discussed. 

\subsection{Distribution Shifts}
\label{sec:assumptions}
Distribution shift \cite{quinonero2008dataset} is a broad topic studying how test data can differ from training data and how would such differences affect model performance. There are several possible causes for dataset shift, out of which two are deemed to be the most important reasons: Sample
selection bias and non-stationary environments according to \cite{DBLP:journals/pr/Moreno-TorresRACH12}. The motivation for referring to notions of data distribution shift in our paper is to utilize measures of distribution shift that provide practical tools as well as rich theoretical backgrounds that enable us to quantify similarity (and/or distance) among pairs of datasets that we will later use for data debiasing. Next, we look at formal definitions and different types of distribution shifts.

If we consider a dataset $(X,Y)$ to be a set of independent and identically distributed \emph{$(i.i.d.)$} set of instances drawn randomly from an unknown continuous probability density function, then a classification problem is defined by a joint distribution \emph{$P(X,Y)$} of features a.k.a. covariates \emph{$X$} and target variables \emph{$Y$} \cite{DBLP:journals/pr/Moreno-TorresRACH12}. According to the Bayesian Decision Theory \cite{DBLP:books/lib/DudaHS01}, a classification can be described either by the prior probabilities of the classes \emph{$P(Y)$} and the class conditional probability density functions \emph{$P(X|Y)$} for all classes \emph{$Y = 1,\dots, c$} where $\emph{c}$ is the number of classes or by the covariate probabilities \emph{$P(X)$} and conditional probability density functions \emph{$P(Y|X)$}. Thus the joint distribution \emph{$P(X,Y)$} can be decomposed in both following forms:
\begin{equation*}
    P(Y|X) = \frac{P(Y)P(X|Y)}{P(X)}, \; \; 
    P(X|Y) = \frac{P(X)P(Y|X)}{P(Y)}
\end{equation*}
where \emph{$P(X)=\sum_{Y=1}^c P(Y)P(X|Y)$} and similarly \emph{$P(Y)=\sum_{X=1}^c P(X)P(Y|X)$} in \emph{$P(Y|X)$} and \emph{$P(X|Y)$} classification problems respectively. The two forms of problem formulation will be formalized as \emph{$Y\rightarrow X$} and \emph{$X\rightarrow Y$} (pronounced as \emph{Y} given \emph{X}) respectively in the rest of this paper.

The literature on distribution shift detection and adaptive learning domain indicates that there are three types of distribution shifts \cite{DBLP:journals/kais/GoldenbergW19, DBLP:journals/tkde/LuLDGGZ19}: 
\begin{enumerate}
    \item \textbf{Covariate shift} appears only in \emph{$X\rightarrow Y$} problems when the probability of input features \emph{$P(X)$} changes, but the decision boundary defining the relationship between covariates and target labels \emph{$P(Y|X)$} remains the same. In other words, the distribution of the input changes, but the conditional probability of a label given an input remains the same. These shifts are known as ``virtual shifts".
    \item \textbf{Prior probability or target shift} appears only in {$Y\rightarrow X$} problems when the probability of target labels \emph{$P(Y)$} changes but \emph{$P(X|Y)$} remains the same. For example, consider the case when the output distribution changes but for a given output, the input distribution stays the same.
    \item \textbf{Concept drift} basically can appear in both types of problems namely in problems of type \emph{$X\rightarrow Y$} where the probability of \emph{$P(Y|X)$} changes between train and test data or in \emph{$Y\rightarrow X$} problems where \emph{$P(X|Y)$} changes. Concept drift happens when the input distribution remains the same between the two datasets but the conditional distribution of the output given an input changes. In other words, the decision boundary defining the relationship between covariates and labels changes.
\end{enumerate}

\subsection{Quantifying Distribution Shift} \label{sec:mmd}
We are interested in measuring distribution discrepancy between two datasets $\bm{X}\in \mathbb{R}^{n \times d}$, and $\bm{Z} \in \mathbb{R}^{m \times d}$ defined over the same feature space of \emph{$d$} features and having an arbitrary size of input samples.
Discrepancy between two datasets can be due to differences in their feature and label distributions, so the conditional probability of labels given an input can remain the same.
Since our goal is to develop a method to cluster similar distributions to enable us to augment a test dataset and also for the sake of maintaining the generality of the problem, we assume we don't have access to target labels and thus do not use prior probability information for shift quantification. As a result, we only use covariate distribution similarities. Moreover, we assume the similarity of the attribute space between different tasks; meaning that all the datasets have the same number of feature with the same range of values. 

One of the most used measures for quantifying pairwise distribution differences is the Kullback-Leibler (KL for short) distance \cite{DBLP:journals/MVN-KLD}. KL has nice theoretical properties, but it is not considered a metric as it is not symmetric ($KL \left(\emph{P, Q}\right)\neq KL  (\emph{Q, P})$ if $\emph{P, Q}$ are probability distributions of covariate sets $\bm{X}$ and $\bm{Z}$ respectively) and it does not satisfy the triangle inequality~\cite{DBLP:journals/kais/GoldenbergW19}. A modified version of KL-divergence which belongs to a symmetrized sub-category of KL-divergence is the Jensen-Shannon divergence \cite{NEURIPS2020_78719f11} which is a metric but still, as with all measures based on KL-divergence, it is sensitive to the sample size and requires both datasets to have the same cardinality.

Another well-known metric for measuring the distance between two distributions is the Maximum Mean Discrepancy (MMD for short) \cite{DBLP:journals/jmlr/GrettonBRSS12_MMD}. It is a multi-variate non-parametric statistic calculating the maximum deviation in the expectation of a function evaluated on each of the random variables, taken over a reproducing kernel Hilbert space (RKHS). MMD can equivalently be written as the L2-norm of the difference between distribution mean feature embeddings in the RKHS. In contrast to KL-Divergence, MMD is not sensitive to the number of instances and can be highly scalable to any arbitrary number of instances for each of the distributions depending on the kernel function employed for its calculation. 

The MMD between two data distributions $\bm{X}\sim P$ and $\bm{Z}\sim Q$ is given by:
\begin{align}\label{eqn:mmd}
    MMD~\bigl(P,Q \bigr) = \big \Vert~\mu_{P} - \mu_{Q}~ \big \Vert^{2} _{\mathcal{H}}
\end{align}
  
Where $\mu_{P}$ is the kernel mean of $\bm{X}$ estimated using $\mu_{P}~(\phi(X)) = \frac{1}{n} \sum_{i=1} \phi(x_{i})$ and similarly $\mu_{Q}$ is kernel mean of $\bm{Z}$ assuming $\phi: \bm{X} \rightarrow \mathcal{H}$ to be a feature map embedding $\bm{X}$ to the embedding Hilbert space $\mathcal{H}$. Then Eq.~\ref{eqn:mmd} can be substituted as: 
\begin{align}\label{eqn:mmd_Fi}
MMD~\bigl(P,Q \bigr) = \Bigg\Vert~{\frac{1}{n} \sum_{i=1} \phi(x_{i}) - \frac{1}{m} \sum_{i=1} \phi(z_{i})}~\Bigg\Vert^{2} _{\mathcal{H}} 
\end{align}

The inner product (indicated by $\langle \, \bullet \, \rangle$ ) of feature means of \(\bm{X} \sim P\) and \(\bm{Z} \sim Q\) can be written in terms of the kernel function such that:
\begin{align}\label{eqn:phi_kernel}
\Bigl \langle~\mu_{P}~\bigl(\phi~\left(\bm{X}~\right)\bigr), \mu_{Q}~\bigl(\phi~\left(\bm{Z}~\right)\bigr)~ \Bigr \rangle_{\mathcal{H}} = E_{P,Q}~\biggl[~\Bigl \langle~ \phi~(\bm{X}~), \phi~(\bm{Z}~) \Bigr \rangle_{\mathcal{H}}~\biggr] = E_{P,Q}~ \Bigl[~k\left(\bm{X, Z}~\right)~\Bigr] 
\end{align}

Substituting Eq.~\ref{eqn:phi_kernel} into Eq.~\ref{eqn:mmd} we can rewrite it such that:
\begin{align}\label{eqn:mmd_kernelize}
MMD~\bigl(P,Q \bigr) = E_{P}~ \Bigl[~k\bigl(\bm{X, X}~\bigr)~\Bigr] - 2 E_{P,Q}~ \Bigl[~k\bigl(\bm{X, Z}~\bigr)~\Bigr] + E_{Q}~ \Bigl[~k\bigl(\bm{Z, Z}~\bigr)~\Bigr] 
\end{align}

Finally, expanding the Eq.~\ref{eqn:mmd_kernelize}, the two sample MMD-test can be calculated by:
\begin{align}\label{eqn:mmd_final}
\begin{split}
    MMD~\bigl(\bm{X, Z}\bigr) = {\frac{1}{n (n-1)} \sum_{i} \sum_{j \neq i} k(x_{i}, x_{j})} - {2 \frac{1}{n.m} \sum_{i} \sum_{j} k(x_{i}, z_{j})} \\ +{\frac{1}{m (m-1)} \sum_{i} \sum_{j \neq i} k(z_{i}, z_{j})}
\end{split}
\end{align}

In \cite{DBLP:journals/jmlr/GrettonBRSS12_MMD} it is suggested to use linear statistic if the datasets are sufficiently large. Since our sample sizes are large enough, we use the linear kernel  for MMD calculations in Eq~\ref{eqn:mmd_final} in our experiments.
To avoid scale differences it is a good practice to normalize the values in the [0,1] range.

\section{Proposed Method}
\label{sec:proposed}
In this section, the detailed procedure of the proposed $\bm{MASC}$ method is described in 4 steps. A procedural overview of each of these steps of the proposed data augmentation method is provided in Algorithm~1. Before approaching these steps with details, an overall overview of the process is discussed in the following.
\RestyleAlgo{ruled}
\LinesNumbered

\SetKwComment{Comment}{$\triangleright$ }{}
\SetKwInput{kwInput}{Input}
\SetKwInput{kwOutput}{Output}

\begin{algorithm}[hbt!] 
\caption{Procedure of the proposed debiasing method \textbf{MASC}}\label{alg:two}

\kwInput{$\bm{D}_{all}$ \textemdash\, Set of all tasks;\leavevmode \newline 
            $b$ \textemdash\, Index of target task $\bm{X}_b$; \Comment*[f]{$\bm{D}_{all}=\bigcup\limits_{i=1}^{r}\bm{X}_{i}$, $b \in \{1,\dots,r\} \Rightarrow \bm{X}_b \subset \bm{D}_{all}$}} 

\kwOutput{$\widehat{\bm{X}}$ \textemdash\, Minority-augmented (debiased) target task}

\For(\Comment*[f]{$r=\big|D_{all}\big|$}){$i\gets0$ \KwTo $r$}
{
    \For{$j\gets 0$ \KwTo $i$}
        {
            $\bm{W}(i,j) \gets MMD(\bm{X}_i,\bm{X}_j)$  \Comment*[r]{Pairwise MMD: Eq.~\ref{eqn:mmd_final}}
                \eIf{$i \neq j$}
            {
                $\bm{A}(w_i,w_j) \gets ~exp\left(-\gamma ~ \big \Vert~w_i-w_j~\big \Vert^{2} ~\right)$ \Comment*[r]{Gaussian Kernel: Eq.~\ref{eqn:rbf}}
            }{
                $\bm{A}(w_i,w_j) \gets 0$ 
                \Comment*[r]{Zeros on diagonals}
            }
        }
}

$\bm{L} \gets \bm{D}- \bm{A}$: where $D_{i, i} \gets \sum_{j=1}^{r}A_{i,j}$; \Comment*[f] Graph Laplacian: Sec~\ref{sec:optimal_k}

$\bm{L} \gets \bm{U}~\Sigma~ \bm{V}^{T}$; \Comment*[f] SVD: Eq~\ref{eqn:svd}

$e \gets \bigl[\lambda_2-\lambda_1, \lambda_3-\lambda_2, \ldots, \lambda_l-\lambda_{l-1} \bigr]$; \Comment*[f] Eigengap vector: Eq~\ref{eqn:eigengap}

$k$ (optimal number of clusters): apply eigen-gap technique on $e$, $k$ is the index of largest gap;

$ \bm{U}\in \mathbb{R}^{r\times k} \gets \begin{bmatrix}
 \vdots & & \vdots \\
 u_1, & \ldots, & u_k\\
 \vdots & & \vdots
\end{bmatrix}
\estimates \min\limits_{1\dots k}\lambda_k $; \Comment*[f] Top $k$ eigenvectors: Eq.~\ref{eqn:clustering}

$\bm{D}_{all} \gets Kmeans(U)$; \Comment*[f] = $Kmeans(U) = \bigl\{\,\bm{C}_1 \cup \ldots \cup \bm{C}_k \,\bigr\}, \quad k\leq r$: Eq.~\ref{eqn:clusters}

$\bm{C}_c \gets \bigl\{~\bm{X}_1 \cup \ldots \cup \bm{X}_t ~\bigr\},\quad t\leq r \quad \& \quad c\in\{~1,\ldots,k~\}$; \Comment*[f] Where $\big|~\bm{C}_c~\big| = \sum_{i=1}^p N_i=~N$: Eq.~\ref{eqn:cluster_c}

\For{$i\gets 1$ \KwTo $p$}
{
    $\big|~ \bm{X}_{~S_i} ~ \big| \gets n_i$; \Comment*[f] Subgroup cardinality of $\bm{X}=\bm{X}_b$, $\sum_{i=1}^p n_i=~n$: Eq.~12
}

$ n_l \gets max~\bigl(~n_1,\ldots,n_p~\bigr) $; \Comment*[f] Cardinality of majority group

\eIf{$N_g~>~n_l$}
{
    $ \widehat{\bm{X}} \gets \bm {X} \cup \bigcup\limits_{j=1}^{(l_g - n_g)} \bm{C}_{S_g}(j)$; \Comment*[f] Augment $\bm{X}\subseteq\bm{C}_c$ in the c-th cluster
}{
    $ \widehat{\bm{X}} \gets \bm{C}_{S_g}$; 
}
   
\end{algorithm}
\label{alg:a1}
Assume having $r$ number of biased datasets $\bm{D}_{all}=\bigl\{~\bm{X}_1 \cup \bm{X}_2 \cup \ldots \cup \bm{X}_r~\bigr\}$ according to $r$ different tasks. For instance, these datasets could each belong to different branches of a franchised hypermarket or be from civil registration offices in different cities (or states) and many other similar cases. Nevertheless, Our final goal is to find a clustering of these datasets based on a similarity score such that in each cluster, tasks can share their instances. This way, an arbitrary dataset $\bm{X}_b \subset \bm{D}_{all}$ that is biased by over-representing a majority\footnote{ (\textit{majority, non-protected}), and also (\textit{minority}, and \textit{protected}) groups will be used interchangeably in this paper.} group w.r.t. a protected attribute, can borrow instances of minority group(s) from its neighboring tasks and construct an augmented unbiased training set. 
For the clustering procedure, a spectral clustering algorithm is utilized that can identify the optimal number of clusters automatically based on an Eigen-gap or Spectral-gap heuristic introduced in \cite{DBLP:journals/sac/Luxburg07}. In order to perform the clustering step, initially we need to construct an \textit{affinity} matrix from the pairwise distances that we obtain by $MMD$ metric.

\subsection{Affinity Matrix Computation} \label{sec:affinity}
The first step in the proposed method is to compute pairwise distance (or discrepancy) between each pair of datasets $\bm{X}_i \subset \bm{D}_{all}$ and $\bm{X}_j \subset \bm{D}_{all}$ using Eq.~\ref{eqn:mmd_final}. The distances are then transformed into a symmetric matrix of pairwise distances $\bm{W}\in \mathbb{R}^{r\times r}$ such that the diagonal of the matrix is all zeros. An intuitive way to convert a pairwise distance matrix into an ''Affinity`` matrix is by applying a \textit{Radial Basis Function a.k.a. Gaussian Kernel} \cite{DBLP:journals/sac/Luxburg07, DBLP:journals/pami/JayasumanaHSLH15}:
\begin{equation}\label{eqn:rbf}
    \bm{A}~\bigl(w_i, w_j \bigr) = 
    \begin{cases}
      ~exp\left(-\gamma ~ \big \Vert~w_i-w_j~\big \Vert^{2} ~\right), & \text{if}\ i~\neq j \\
      ~0, & \text{otherwise}
    \end{cases}
\end{equation}
where $w_i~\text{and}~w_j$ are two entries of the distance matrix $\bm{W}$. Eq.~\ref{eqn:rbf} results in a weighted undirected symmetric affinity matrix $\bm{A}$ with zero diagonal elements with weights being Gaussian functions of the pairwise distances.

\subsection{The Optimal Number of Clusters k}\label{sec:optimal_k}
In order to perform spectral clustering on the affinity matrix, we need to calculate the unnormalized graph Laplacian \cite{DBLP:journals/sac/Luxburg07} {$\bm{L} = \bm{D}- \bm{A}$} where $D_{i, i}=\sum_{j=1}^{r}A_{i,j}$ is the diagonal degree matrix of the affinity matrix $\bm{A}$. Graph Laplacian is key to spectral clustering; its eigenvalues and eigenvectors reveal many properties about the structure of a graph. 

According to the ''Perturbation Theory``, an optimal number of clusters \textit{k} for a dataset can be given through the eigengap identification of eigenvalues of the graph Laplacian, which is the largest difference between eigenvalues \cite{DBLP:journals/jmlr/LittleMM20, DBLP:journals/nca/NatalianiY19}. Thus, computing the eigenvalues of the Laplacian matrix and finding it's biggest gap can discover the optimal number of clusters. This way, one can avoid the difficult and tricky decision of the cluster number parameter. Thus, similar to the instructions in \textit{step 5} of \cite{DBLP:journals/spl/ParkHKN20} we perform a ''\textit{Singular Value Decomposition (SVD)}`` to calculate the eigenvalues of the Laplacian matrix $\bm{L}$: 
\begin{align}\label{eqn:svd}
    \bm{L} &= \bm{U}~\Sigma~ \bm{V}^{T}
\end{align}
where $\bm{U, V}$ are unitary matrices called left and right singular matrices, respectively containing eigenvectors corresponding to eigenvalues in $\Sigma$. Next, we create an eigengap vector $e$ using the eigenvalues from $\Sigma$ in Eq.~\ref{eqn:svd} as follows:
\begin{align}\label{eqn:eigengap}
    e = \bigl[\lambda_2-\lambda_1, \lambda_3-\lambda_2, \ldots, \lambda_l-\lambda_{l-1} \bigr]
\end{align}
where $\lambda_k$, is the \textit{k-th} sorted eigenvalue in ascending order. Note that, if $\bigl(\lambda_k - \lambda_{k-1}\bigr)$ implies the largest difference i.e. eigengap according to Eq.~\ref{eqn:eigengap}, then index $\textit{k}$ is the optimal number of clusters.

\subsection{Spectral Clustering} \label{sec:spectral}
After obtaining \textit{k}, the desired number of clusters in Section~\ref{sec:optimal_k}, there is one more step to finally be able to partition the affinity matrix. In this step, we find the top k eigenvectors $u_1,\ldots, u_k$ according to the top k smallest eigenvalues of the Laplacian, stack them as columns of a new matrix $\bm{U}\in \mathbb{R}^{r\times k}$ such that:
\begin{align} \label{eqn:clustering}
  \bm{U} = 
\begin{bmatrix*}
 \vdots & & \vdots \\
 u_1, & \ldots, & u_k\\
 \vdots & & \vdots
\end{bmatrix*}
\estimates \min\limits_{1\dots k}\lambda_k 
\end{align}
where $\estimates$ stands for the term ''Corresponding to``. Then, a k-means clustering~\cite{DBLP:journals/pr/LikasVV03} is performed on the rows of matrix $\bm{U}$ which is equivalent to a clustering of the r datasets:
\begin{align} \label{eqn:clusters}
   \bm{C} = Kmeans(U) = \bigl\{\,\bm{C}_1 \cup \bm{C}_2 \cup \ldots \cup \bm{C}_k \,\bigr\} \quad \text{and} \quad k\leq r \quad \bm{C} \equiv \bm{D}_{all}  
\end{align}
 and $\equiv$ sign represents the equivalence of its operands. Note that, in practice, spectral clustering is often followed by another clustering algorithm such as k-means to finalize the clustering task. The main property of spectral clustering is to transform the representations of the data points of $\bm{X}_b$ into the indicator space in which the cluster characteristics become more prominent and passes much more processed/meaningful information to the next step clustering algorithm.

\subsection{Data Augmentation Within Clusters} \label{sec:augment}
Now that the set of input tasks/datasets $\bm{D}_{all}$ is clustered into $k$ partitions according to Eq.~\ref{eqn:clusters}, the data augmentation process for minority group(s) can be fulfilled. If cluster $c$ consists of $t$ datasets:
\begin{align} \label{eqn:cluster_c}
    \bm{C}_c = \bigl\{~\bm{X}_1 \cup \ldots \cup \bm{X}_t ~\bigr\} \quad \text{where}\quad t\leq r \quad \& \quad c\in\{~1,\ldots,k~\}
\end{align}
Initially, we create a pool of instances in the cluster $\bm{C}_c$ by collecting all the instances from each dataset belonging to the cluster. The number of instances in this cluster $\big|~\bm{C}_c~\big| = N$, (where $\big|\, \bullet \,\big|$ denotes cardinality) can be written as a sum of the number of instances belonging to each of the $p$ protected groups $\sum_{i=1}^p N_i=~N$. Given a protected attribute $S=\left\{~S_1, \ldots, S_p~\right\}$ with $p$ groups and knowing $\big|~\bm{X}~\big| = n$, the augmentation process for task $\bm{X} \subset \bm{D}_{all}$ is a very straightforward process based on protected groups cardinality. We calculate the cardinality of each group corresponding to the number of instances belonging to that group such that:
\begin{align}
    \Big|~ \bm{X}_{~S_i} ~ \Big| = n_i \quad \textrm{for i}~\in \big\{~1,\ldots,p~\big\}
\end{align}\label{eqn:cardinality}
where $\sum_{i=1}^p n_i=~n$. Next, we identify the biggest group and indicate it as the majority/non-protected group through a procedure like $max~\bigl(~n_1,\ldots,n_p~\bigr) = n_l $. Ideally, the intention would be to balance every minority subgroup $g$ to have a cardinality as big as the majority group $l$, so that $\big|~\widehat{\bm{X}}_{~S_g} ~\big| = n_l$. Thus, every protected group needs to be augmented by a difference of $n_l -n_g$. However, it is only the case if the pool of shared protected group instances includes this number of instances otherwise we augment by as many instances as there exist in the shared pool. Thus, the augmented version of dataset $\bm{X}$ has the following number of instances depending on the number of shared instances:
\begin{align}
    \widehat{\bm{X}}~ \leftarrow
    \begin{cases}
       \; \bm {X} \cup \bigcup\limits_{j=1}^{(n_l - n_g)} \bm{C}_{S_g}(j) \quad & \textrm{if}\; N_g~>~n_l  \\
       \; \bm{C}_{S_g} & \text{otherwise}
    \end{cases}
\end{align}\label{eqn:x_new}
where $\bm{C}_{S_g} = \bigl\{~\bm{C}_c~\lvert~S=S_g ~\bigr\}$. Note that, $\bm{C}_c$ and $\bm{X}_b$ are substituted by $\bm{C}$ and $\bm{X}$ respectively, to avoid syntax complication.

\section{Experimental Results}
\label{sec:EXP}
In order to evaluate the effectiveness of the proposed \emph{MASC} method, in this section the conducted experimental results on a number of real-world datasets are analyzed. The organization of this section is as follows: First, details of the datasets employed are presented. Next, the evaluation measures used in the experiments and also methods used for comparisons are described. Finally, the experimental results and discussions on them are provided.

\subsection{Datasets}
To evaluate \textit{MASC}'s performance in addressing group imbalance and representation bias, we used the recently released US Census datasets \cite{DBLP:conf/nips/DingHMS21}, which comprise a reconstruction of the popular Adult dataset \cite{Dua:2019}. These datasets provide a suitable benchmark with 52 datasets representing different states, effectively capturing the problem of group imbalance between states with varying numbers of instances but similar feature spaces.

The datasets \cite{DBLP:conf/nips/DingHMS21} include census information on demographics, economics, and working status of US citizens. Spanning over 20 years, they allow research on temporal and spatial distribution shifts and incorporate various sources of statistical bias. As already mentioned in Section~\ref{sec:mmd}, in this study we assume that the conditional probability of labels given specific inputs remains constant. Therefore, we focus on the latest release, specifically the year \emph{2019} (till the date of submission), and examine the spatial context to explore the connection between covariate shifts and bias.

The feature space consists of 286 features, with only 10 deemed relevant \cite{DBLP:conf/nips/DingHMS21}. The target variable, \emph{Income Value}, is transformed into a binary vector to predict whether an individual earns an income of more than 50k:  $Income \in  \{\le50K, >50K\}$, the positive class being $``>50K''$. We selected \emph{``Race''} as the protected attribute due to the challenge it poses compared to gender or age, given the highly imbalanced distribution of racial groups across states. The \emph{``Race''} attribute has 9 categories, but due to a very small representation of seven of these categories which usually comprise less than 1\% of the instances in the dataset, we aggregate them to a bigger group called \emph{``Other''}. Thus, the categories in our experiments are aggregated into 3 groups: \textit{White, Black, and Other}. Categorical features are transformed into numerical features and all the features are normalized by standard scaling using their mean ($\mu$) and standard deviation ($\sigma$) values such that each $z=(x-\mu)/\sigma$ is a standard representation of its $x$ and lies within the range $[0,1]$.

\begin{table}[t]
\small
\centering
\caption{Statistics for five chosen states including their Racial group imbalance ratio, Abbreviations denoted as (Abbr), and Class Ratio for positive~(Pos) and Negative~(Neg) classes.}

\label{tab:dataset_details}
\setlength{\tabcolsep}{8.5pt} 
\begin{tabularx}{1.0\linewidth}{lccccccc}
    \toprule
    &&&&
    \multicolumn{3}{c}{\bf{Group Distribution Ratio}} &\bf{Class Ratio}\\
    \cmidrule(r){5-7}
    \bf{States} & 
    \bf{Abbr} &  
    \bf{Samples} &
    \bf{Cleaned} & 
    \bf{White} &
    \bf{Black} &
    \bf{Other} &
    \bf{(Pos|Neg)} \\
    \midrule
      Colorado       & CO & 57,142 & 32,264   & \bf{87.98\%}  & 2.56\%   & 9.46\%    & 1:1.26  \\ 
    
    North-Dakota          & ND  &   7,960 &  4,455  & \bf{92.03\%}  & 1.35\%   & 6.76\%    & 1:2.2  \\

    Maryland       & MD &  60,237 &  32,865  & \bf{64.59\%}  & 22.4\%   & 13.01\%   & 1.05:1  \\

    Mississippi    & MS  &  29,217 &  13,159  & \bf{66.43\%}  & 29.92\%  & 3.66\%   & 1:2.51  \\
    
    Montana          & MT  &  10,649 &  5,547  & \bf{92.03\%}  & 0.36\%   & 7.61\%   & 1:2.09  \\
\bottomrule
\end{tabularx}
\end{table}


Refer to Table~\ref{tab:dataset_details} for detailed information on the filtered (cleaned) datasets, including racial distribution, class imbalance ratio, name abbreviation conventions, and other details. The table summarizes information for 5 out of 51 datasets. The intuition behind this specific selection of states will be addressed in detail in Section~\ref{sec:results}. The datasets exhibit significant racial bias, with the \textit{White} group representing the majority (also referred to as non-protected) in all 5 datasets.

\subsection{Metrics}
\label{sec:bias measure}

In this paper, we adopt five measures in total. We use accuracy~\cite{DBLP:conf/icse/VermaR18} for analyzing models predictive performance along with four measures for bias and fairness quantification; \textit{Disparate Impact}~\cite{DBLP:conf/kdd/FeldmanFMSV15}, \textit{Statistical (or Demographic) Parity}~\cite{DBLP:journals/corr/zoo}, \textit{Equalized Odds}~\cite{DBLP:conf/icse/VermaR18}, and a new proportionality metric that we introduce, the \textit{Group Distribution Ratio} for quantification of bias on datasets before and after debiasing. The measures that take into account model outcome or in other words, which involve model training and prediction (e.g. Accuracy and Equalized Odds) are not relevant for the first part of experiments. Given a dataset $\bm{X}=\bigl\{D, S, Y ~\bigr\}$, with $D$ regular features, a protected feature $S$ (i.e. \textit{Race}) and a binary target class, the disparate impact (DI short for) of the given dataset is calculated as follows:
\begin{align} \label{eqn:Disp Imp}
    DI = \frac{P\,\bigl(\,Y=1 \,\big| \,S=0\,\bigr)}{P\,\bigl(\,Y=1\, \big| \,S=1\,\bigr)}
\end{align}
which basically calculates the ratio of the probability of being a member of the protected group having positive outcomes to the probability of the non-protected group with positive outcomes. DI ranges between zero and one $DI\in(0,2)$ with $1$ being the best value i.e. implies there is no bias. $0, 2$ mean maximum bias toward one group or the other respectively.

The measure statistical parity (SP for short) also computes a quite similar value, where it reflects the mentioned change as a difference instead of a ratio:
\begin{align} \label{eqn:Stat Par}
    SP = P\,\bigl(\,Y=1 \,\big| \,S=0\,\bigr) - P\,\bigl(\,Y=1\, \big| \,S=1\,\bigr)
\end{align}
Since we consider two protected groups of ''\textit{Black}`` and ''\textit{Other}`` in our experiments, the results are calculated for each of the measures twice; for each of the two protected groups against the non-protected group of ''White``. So in our analysis $S\in\{~0,1~,2~\}$. SP takes values in the range $SP\in(-1,1)$ with 0 as the best possible value implying zero bias.

The measure \textit{Equalized Odds} (Eq.Odds for short) calculates the difference in prediction errors between the protected and non-protected groups for both classes as $\left|{\delta}FPR \right| + \left| {\delta}FNR\right|$ where ${\delta}FNR$ stands for ''\textit{False Negative Rates}`` and ${\delta}FPR$ stands for ''\textit{False Positive Rates}`` that are also known as \emph{Equal Opportunity} and \textit{Predictive Equality} respectively. The ${\delta}FNR$ measures the difference of the probability of subjects from both the protected and non-protected groups that belong to the \textbf{positive} class to have a negative predictive value and similarly, the ${\delta}FPR$ calculates the difference of the probability of subjects from both the protected and non-protected groups that belong to the \textbf{negative} class to have a positive predictive value. So, the Eq.Odds is formulated as follows:
\begin{align} \label{eqn:EqOdds}
    Eq.Odds = \left| {P(\hat{Y} = 0|Y = 1, g = w) - P(\hat{Y}=0|Y = 1, g = b/o)}\right| +\\
    \left| {P(\hat{Y}= 1|Y = 0, g = w) - P(\hat{Y} = 1|Y = 0, g = b/o)}\right|
\end{align}
where $\hat{Y}$ is the predicted label, Y is the actual label and $g \in G = \{w, b, o\}$ is the protected attribute. The value range for each of ${\delta}FNR$ and ${\delta}FPR$ is [0,1], where 0 stands for a classifier satisfying perfectly the measure with no discrimination and 1 stands for maximum discrimination. Thus, Eq.Odds can range between [0,2]. In this study, $w$ is taken as the majority (non-protected) group and $b$ and $o$ are minority (protected) groups.  

Finally, we introduce a group-proportional measure: the group distribution ratio (GR for short) essentially calculates group imbalance or the proportion of instances belonging to each of the protected groups or the non-protected group w.r.t. the total number of instances in the dataset. Similar to the definition of protected attribute and its member groups in Section~\ref{sec:augment}, the group distribution ratio for a protected group $g$ is obtained as follows: 
\begin{align} \label{eqn:Imbalance}
    GR_g = P\,\bigl(\,X\,\big| \,S=S_g\,\bigr) = \frac{\big|~X_{S_g}~\big|}{\big|~X~\big|} = \frac{n_g}{\sum_{i=1}^{p}n_p}
\end{align}
where the denominator of the fraction in Eq.~\ref{eqn:Imbalance}, is the sum of the cardinality of all subgroups of task $X$ or the total number of its instances $\sum_{i=1}^{p}n_p = n$. Clearly, the cumulative probability of all subgroups $\sum_i^p P\,\bigl(\,X\, \big| \,S = S_i\,\bigr)$ is $=~1$. Thus, a dataset is group balanced w.r.t. a protected attribute if Eq.~\ref{eqn:Imbalance} is proportionately equal for each subgroup. In other words, the optimal balance for each subgroup in the dataset is given by $GR^{\ast} = 1/(\sum_{i=1}^p S_i)$ which implies balanced groups with the same number of instances. As a result, for a protected attribute with two subgroups, the optimal group distribution ratio would be $GR^{\ast}=1/2$ and similarly for a protected attribute with three subgroups $GR^{\ast}=1/3$.

\subsection{Competitors} \label{sec:competitor}
In order to compare the results of our \textit{MASC} augmentation method, we compare it with 4 different competitors including the original shape of untouched datasets along with three other strategies. Specifically, we use a variation of \textit{SMOTE}~\cite{DBLP:journals/jair/0001GHC18} for synthetic minority protected group over-sampling instead of minority class augmenting, and similarly we use a variation of \textit{RUS}~\cite{mishra2017handling}, as a random group under-sampling. In addition, we also introduce a natural geographical neighborhood augmentation by concatenating datasets within their local clusters of geographical neighbors based on the formal region categorization as in~\cite{su6063915}. All the augmentation methods are also analyzed by feeding their outputs to a Logistic Regression classifier (LR for short).

Note that we implement a variation of SMOTE and RUS to over/under-sample based on the protected group distribution of the protected attribute such that: for SMOTE we over-sample both minority groups until their cardinality is as large as the majority group. For the RUS method, we under-sample the majority group and the bigger minority group until they contain as few samples as the smallest minority group. 

\subsection{Empirical Results}
\label{sec:results}
The forthcoming experiments in this section are conducted in order to compare the initial dataset biases of the original datasets before and after the proposed data augmentation and also in comparison with the three other augmentation strategies, respectively  as mentioned in section~\ref{sec:competitor} based on the introduced measures in Section~\ref{sec:bias measure}. Following that, we also compare predictive performance and fairness of a LR classifier on the different augmentation strategies to see how would each of the augmentation methods affect model performance\footnote{The source code of the proposed \textit{MASC} and the comparisons can be found at: \href{https://github.com/SiamakGhodsi/MASC.git}{Github/SiamakGhodsi/MASC}}. Meanwhile, to get an intuition about the step-wise procedure of the proposed \textit{MASC} method, a demonstration of implementation on the aforementioned US-Census datasets, following the steps in Section~\ref{sec:proposed} is illustrated before discussing performance results.  

\subsubsection{A Demo implementation}
\begin{figure}[!htbp]%
    \centering
    \begin{subfigure}[b]{0.29\textwidth}
       \includegraphics[width=1\linewidth]{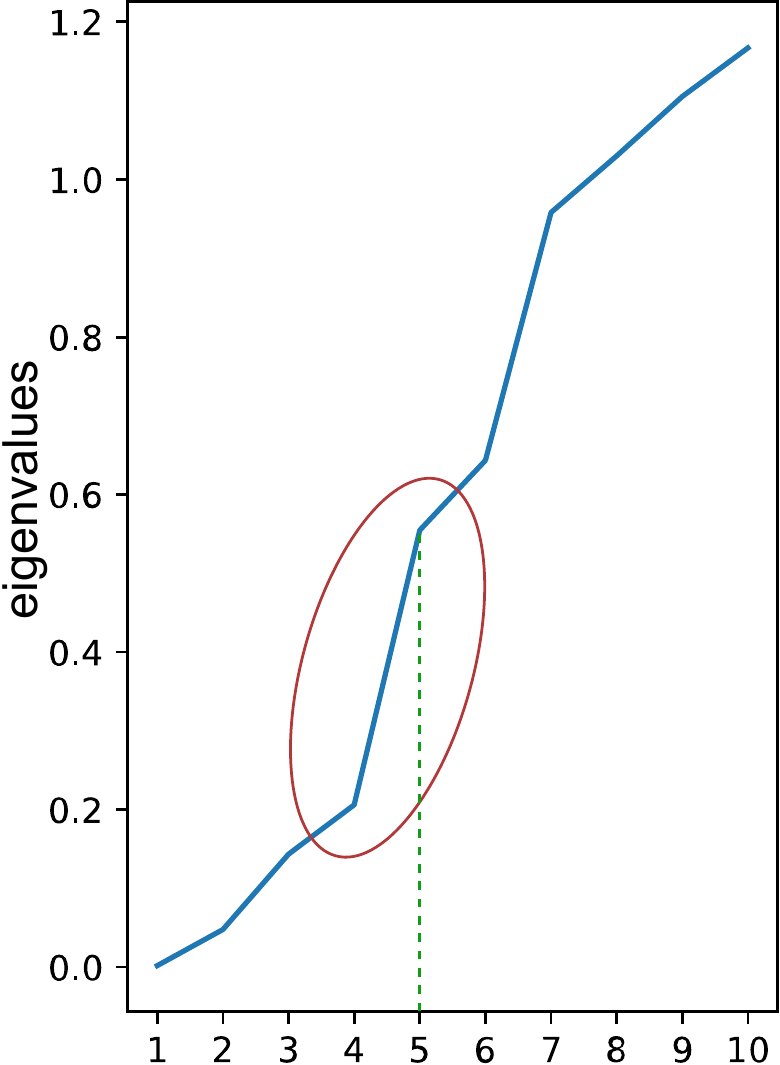}
       \caption{The largest eigen-gap}
       \label{fig:eigengap} 
    \end{subfigure}
    \hspace{0.02\textwidth}   
    \begin{subfigure}[b]{0.66\textwidth}
       \includegraphics[width=1\linewidth]{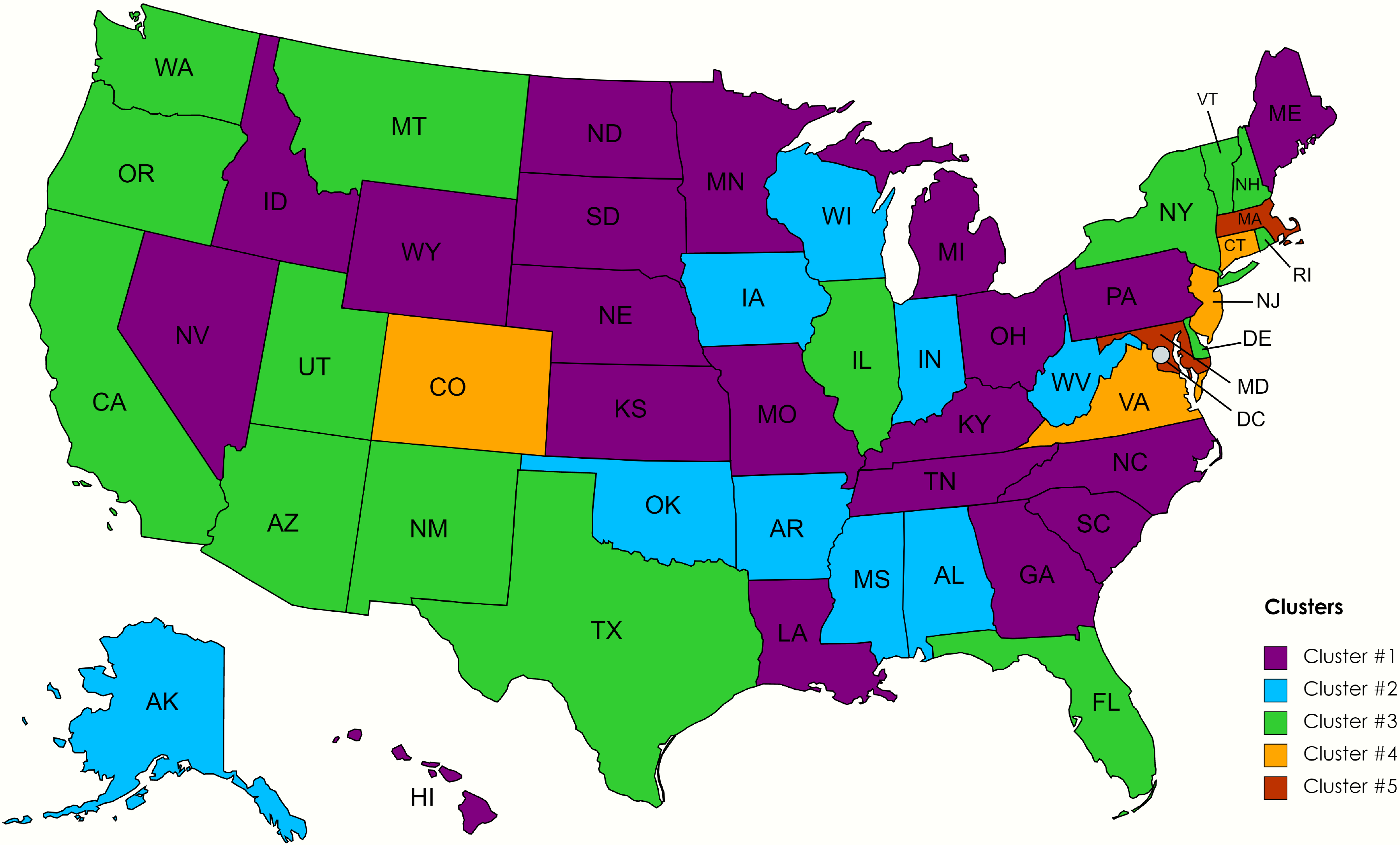}
       \caption{States data clustering shown as a 5-nn graph for legibility}
       \label{fig:spectral}
    \end{subfigure}
    \label{fig:demo}
    \caption{In \textbf{(a)} the first 10 eigenvalues of the Graph Laplacian are shown. The largest eigengap indicating the optimal number of clusters according to Eq.~\ref{eqn:eigengap} is identified between eigenvalues $eig_4$ and $eig_5$ implying that 5 clusters can best partition the graph affinity matrix defined in Section~\ref{sec:affinity}. In \textbf{(b)} Clustering result of the graph Laplacian according to the \textit{Unnormalized Spectral Clustering} method introduced in Section~\ref{sec:spectral} to the optimal number of 5 clusters obtained in the previous step. States are labeled with 2-letter standard US state abbreviations and are coloured based on the clusters they belong to.}
\end{figure}

Initially, an affinity matrix according to steps~1-9 of Algorithm~1 is generated. Following that, based on instructions in lines~11-12 of Algorithm~1, initially a graph Laplacian and afterward its SVD decomposition are calculated from the affinity matrix, in order to obtain eigenvalues of the Laplacian and find the spectral eigengap as in steps~13-14. According to the spectral graph theory \cite{stewart1990matrix}, in an ultimately well-shaped problem, one can observe that there exists an ideal case of k completely disconnected components which constitute a block diagonal Laplacian matrix that has k zero eigenvalues and corresponding k eigenvectors of ones. In this extreme case, the (k+1)-th eigenvector which is non-zero, has a strict gap. This gap identifies the optimal number of connected components that can be clustered as highly similar objects. The eigengap heuristic is an advanced guide to avoid parameter selection, although our problem as well as the majority of real-world problems do not produce such a well-formed block diagonal Laplacian. In Figure~\ref{fig:eigengap} the first ten eigenvalues and the major eigengap are demonstrated. It depicts that our datasets can be semi-optimally clustered into five categories.

We partition the obtained Affinity matrix into five clusters following instructions in Section~\ref{sec:spectral} and accordingly steps~16-17 of Algorithm~\ref{alg:a1}. The clustering is illustrated in Figure~\ref{fig:spectral}. Each color represents a cluster.

Next, according to steps in Section~\ref{sec:augment} and steps~18-26, we augment each of the input datasets using the shared protected-group instances from their neighbors in the same clusters. The results of the minority group(s) augmentation are summarized in Table~\ref{tab:augment} and group distribution and GR scores are shown in Figure~\ref{fig:dist_augment}.

\subsubsection{Results}
\label{results}
The \textit{MASC} is applied to all the input datasets and augments each task based on the cluster that they belong to and therefore the neighborhood instances that they share. Since the method indicates 5 clusters as depicted in Figure~\ref{fig:spectral}, for the sake of readability we evaluate the results for 5 of the datasets, each chosen from one of the clusters. Namely, the states are; \textit{Montana} (MT), \textit{Mississippi} (MS), \textit{North-Dakota} (ND), Colorado (CO), and Maryland (MD). The selection of states within each cluster is based on diversity; we chose states from western and more central regions to northern and east-most states that allow comparing population texture of different regions of the US according to \cite{su6063915}.

\begin{figure}[!htbp]%
    \centering
    \begin{subfigure}[b]{1\textwidth}
       \includegraphics[width=1\linewidth]{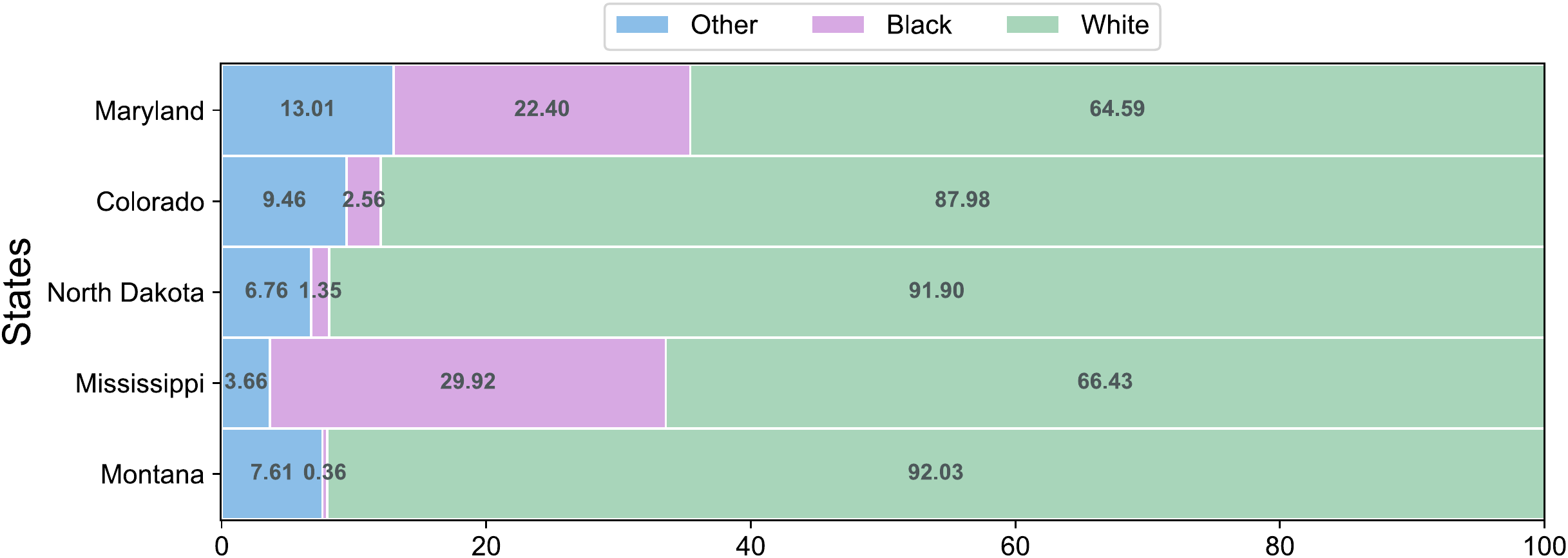}
       \caption{}
       \label{fig:dist_original} 
    \end{subfigure}

    \begin{subfigure}[b]{1\textwidth}
       \includegraphics[width=1\linewidth]{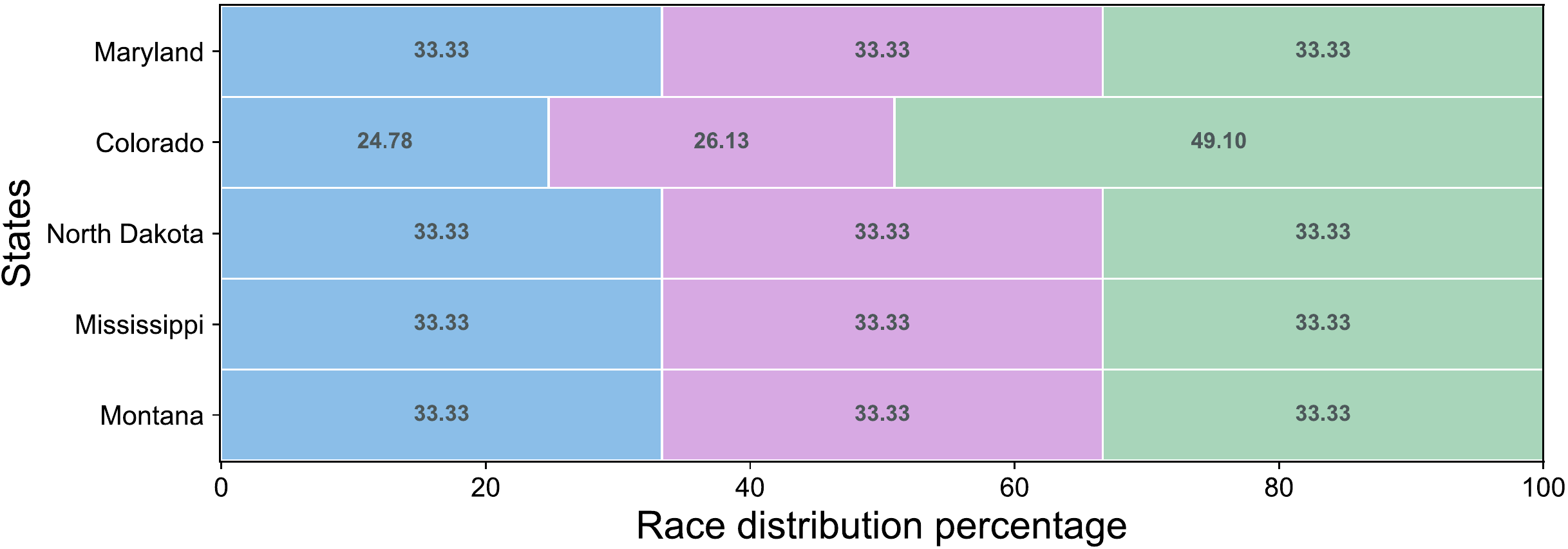}
       \caption{}
       \label{fig:dist_augment}
    \end{subfigure}
    \caption{Comparison of groups distribution ratio (GR) percentage between the original datasets before data augmentation and their augmented versions after applying \textit{MASC} w.r.t. \textit{Race} as protected attribute. categories are: \textit{\{White, Black, Other\}} per state.}
    
\end{figure}

In Figure~\ref{fig:dist_original} the GR values according to the distribution of each of the original datasets is shown. Comparing them to distributions in Figure~\ref{fig:dist_augment} which is the results obtained by our method, the performance of our method in reducing the group differences is inferable. The proposed method borrows similar instances for each minority group from similar states and balances exactly perfectly four of the states and to a very good extent also the Colorado state. In Colorado's case, the number of minority group instances borrowed from other states in the cluster is not enough to equalize the representation of minority groups, but still decreases the worst imbalance in the original dataset in terms of difference between GR-values of Maj-min1 from $87.98\%-2.56\%\approx85\%$ to $49.16\%-26.13\%\approx23\%$ and reduces the $85\%$ difference to $23\%$.   

\begin{table}[t]
\small
\centering
\caption{Evaluation results of the five datasets: the initial stats of the original data in comparison to 4 different augmentation strategies w.r.t. Group Distribution Ratio (GR), Statistical Parity, and Disparate Impact measures. Maj, Min1, and Min2 stand for the Majority (White) and the two Minority (Black, and Other) groups respectively. Boldfaced values imply better results in SP and DI measures. For GR results, the balancedness of the three groups implies better results.}

\label{tab:augment}
\setlength{\tabcolsep}{5pt} 
\begin{tabularx}{\linewidth}{lcccccccc}
    \toprule
     \multirow{2}{*}{\bf{States}}&  \multirow{2}{*}{\bf{Method}}& \multicolumn{3}{c}{\bf{Group Distribution Ratio}} &
    \multicolumn{2}{c}{\bf{Statistical Parity}~(SP)} &
    \multicolumn{2}{c}{\bf{Disparate Impact}~(DI)}\\
    
    \cmidrule(lr){3-5}
    \cmidrule(lr){6-7}
    \cmidrule(lr){8-9}
    
     & 
     & 
    \bf{Maj} & 
    \bf{Min1} & 
    \bf{Min2} & 
    \bf{Min1} &
    \bf{Min2} &
    \bf{Min1} &
    \bf{Min2}\\
    \midrule
    \multirow{5}{*}{\bf{Montana}} &
   Initial &  92.03\%  &  0.36\% &  7.60\%  &    -0.124    &   -0.122  &  0.617   &    0.634 \\

   & MASC & 33.33\%  &  33.33\% &  33.33\%  &    -0.054    &   0.093  &  0.855   &   1.285\\

    & Geo-nei & 83.35\%  &  01.89\% &  14.75\%  &    -0.112    &   -0.070  &  0.726   &   0.832\\
   
    & SMOTE & 33.33\%  &  33.33\% &  33.33\%  &    \textbf{-0.002}    &   \textbf{-0.015}  &  \textbf{0.991}   &    \textbf{0.954}\\

    & RUS & 33.33\%  &  33.33\% &  33.33\%  &    0.075    &   -0.150  &  1.272   &    0.571\\

    \addlinespace
    \multirow{5}{*}{\bf{Mississippi}} & 
     Initial   &  66.42\%  &   29.91\%    &  3.65\%     &   -0.172   &  -0.056   &  0.487  &   0.803  \\

     &MASC   &  33.33\%  &   33.33\%    &   33.33\%     &   -0.072    &  0.094  &   0.782  &   1.297  \\

     & Geo-nei & 80.98\%  &  14.83\% &  04.17\%  &    -0.154    &   \textbf{-0.062}  &  0.559   &    0.810\\
    
    & SMOTE & 33.33\%  &  33.33\% &  33.33\%  & -0.092    &   -0.006  &  0.705   &    \textbf{1.022}\\

    & RUS & 33.33\%  &  33.33\% &  33.33\%  &  \textbf{0.026}    &   0.004  &  \textbf{1.093}   &    1.014\\

   \addlinespace
   \multirow{5}{*}{\bf{North-Dakota}} & 
    Initial   & 91.89\%  &  1.34\%   &   6.75\%  & -0.282  &   -0.181   &    0.261     &   0.535 \\ 
    
    &MASC  &  33.33\%  &  33.33\%   &    33.33\% & -0.098  &  -0.077   &   0.762   &   1.216 \\ 

    & Geo-nei & 90.74\%  &  03.60\% &  5.65\%  &   -0.146   &   -0.112  &  0.598   &   0.691\\
    
    & SMOTE & 31.31\%  &  34.34\% &  34.34\%  &  \textbf{-0.005}    &   \textbf{-0.020}  &  \textbf{0.984}   &    \textbf{0.945}\\

    & RUS & 31.31\%  &  34.34\% &  34.34\%  &    -0.058    &   -0.058  &  0.820  &    0.820\\

     \addlinespace
     \multirow{5}{*}{\bf{Colorado}} & 
     Initial  &  87.98\%  &  2.56\%   &    9.46\%  & -0.176  &  -0.097   &  0.605     &  0.783 \\

    &MASC  &  49.09\%  &  26.12\%   &    24.77\%  & -0.199  &   -0.100   &   0.492     &   0.723\\

    & Geo-nei & 83.02\%  &  02.87\% &  14.10\%  &    -0.125  &  -0.128  &  0.668   &  0.672\\
    
    & SMOTE & 33.33\%  &  33.33\% &  33.33\%  &    -0.007  &  -0.016  &  0.983   &    0.963\\

    & RUS & 33.33\%  &  33.33\% &  33.33\%  &    \textbf{-0.005}  &  \textbf{-0.012} &  \textbf{1.012} &   \textbf{0.971}\\

    \addlinespace
    \multirow{5}{*}{\bf{Maryland}} &
     Initial &   64.58\%   &  22.4\%    &    13.0\% &    -0.108 & -0.082  & 0.798    &   0.842  \\

    &MASC  &   33.33\%   &  33.33\%    &    33.33\% &    -0.101 & 0.060  & 0.82   &  0.931    \\

    & Geo-nei & 74.82\%  &  16.15\% &  09.02\%  &    -0.147   &   -0.061  &  0.64   &   0.871\\
    
    & SMOTE & 33.33\%  &  33.33\% &  33.33\%  &  -0.060    &   0.006  &  0.886   &    \textbf{0.988}\\

    & RUS & 33.33\%  &  33.33\% &  33.33\%  &    \textbf{0.001}    &   \textbf{0.001} &  \textbf{0.996}   &    0.996\\
    
\bottomrule
\end{tabularx}
\end{table}

\label{tab:augmented}
Table~\ref{tab:augment} summarizes the evaluation of the five datasets based on previously introduced measures DI, SP, GR (refer to Section~\ref{sec:bias measure} for details) comparing the results of original states with \textit{MASC}, Geographical-neighborhood grouping, the SMOTE, and the RUS methods. Note that the Geographical-neighborhood augmentation is abbreviated as Geo-nei in the table. The Maj, Min1, and Min2 notations correspond to Majority (\textit{White}) and two minority groups (\textit{Black} and \textit{Other}), respectively. It is empirically shown in the table that the results of the MASC, alleviate group imbalance to a good extent for all the datasets according to the GR column and subsequently achieve good DI and SP rates compared to the original datasets. Moreover, in comparison to Geo-nei, our method performs better for all the states except for the Min1 group in the Colorado state and still achieves much better balances for the protected groups but only the class distribution is slightly worse. Compared to SMOTE and RUS, our method performs comparably well in terms of GR-values but w.r.t. DI and SP metrics, the method reports slightly worse results that is because our method only balances out group distributions and doesn't take into account the distribution of target-class. However, in model performance, our method outperforms the SMOTE and RUS for all the states w.r.t. accuracy and eq.odds metrics that we will see in the followings. Also, there are some technical issues/limitations that may arise using the SMOTE and RUS methods which will be discussed more detail in Section~\ref{sec:discussion}.

\begin{figure}[!htbp]%
    \centering
    \begin{subfigure}[b]{0.48\textwidth}
       \includegraphics[width=1\linewidth]{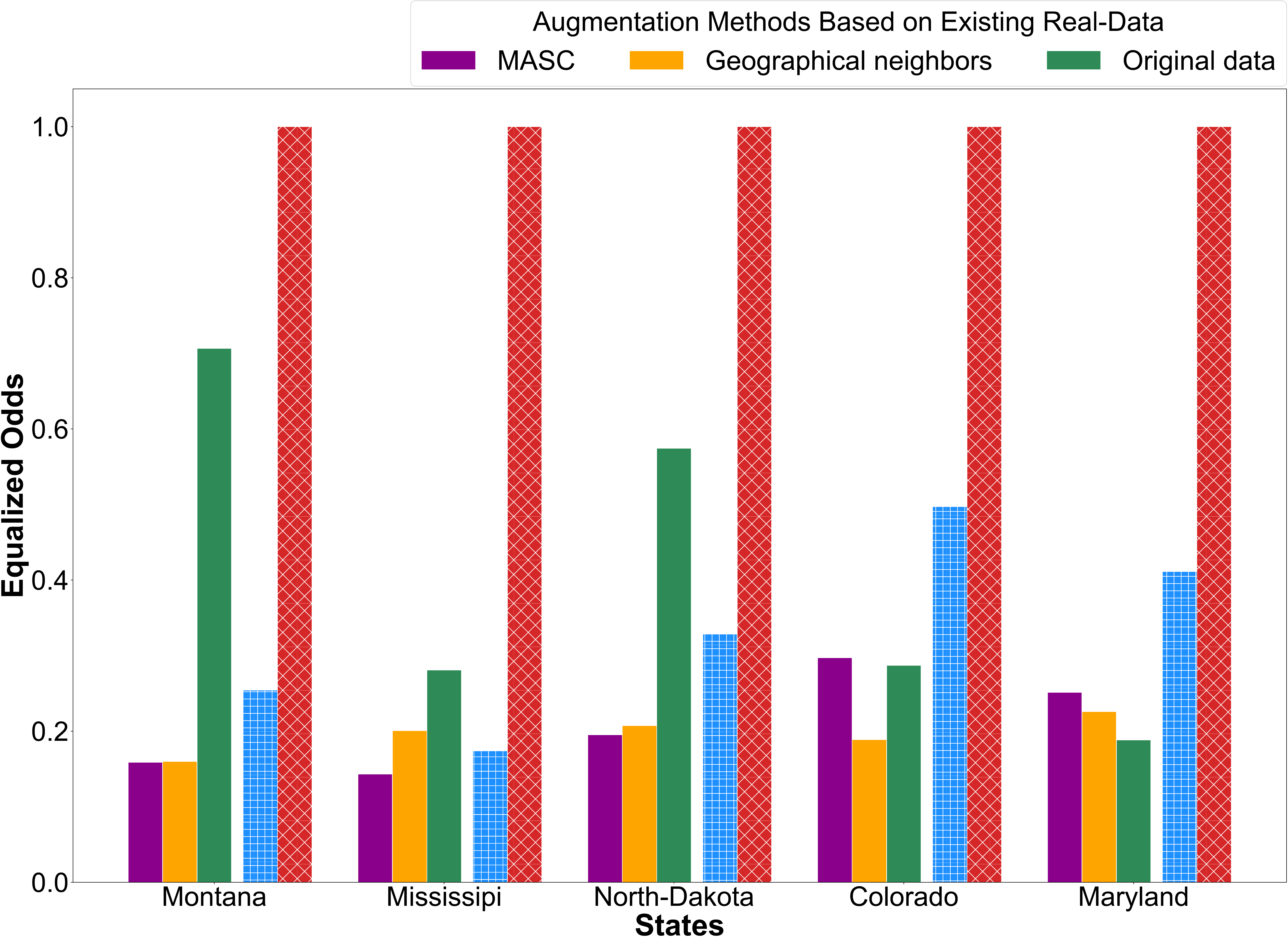}
       \caption{Equalized odds of each method on each of the five states}
       \label{fig:fair} 
    \end{subfigure}
    \hspace{0.02\textwidth}   
    \begin{subfigure}[b]{0.48\textwidth}
       \includegraphics[width=1\linewidth]{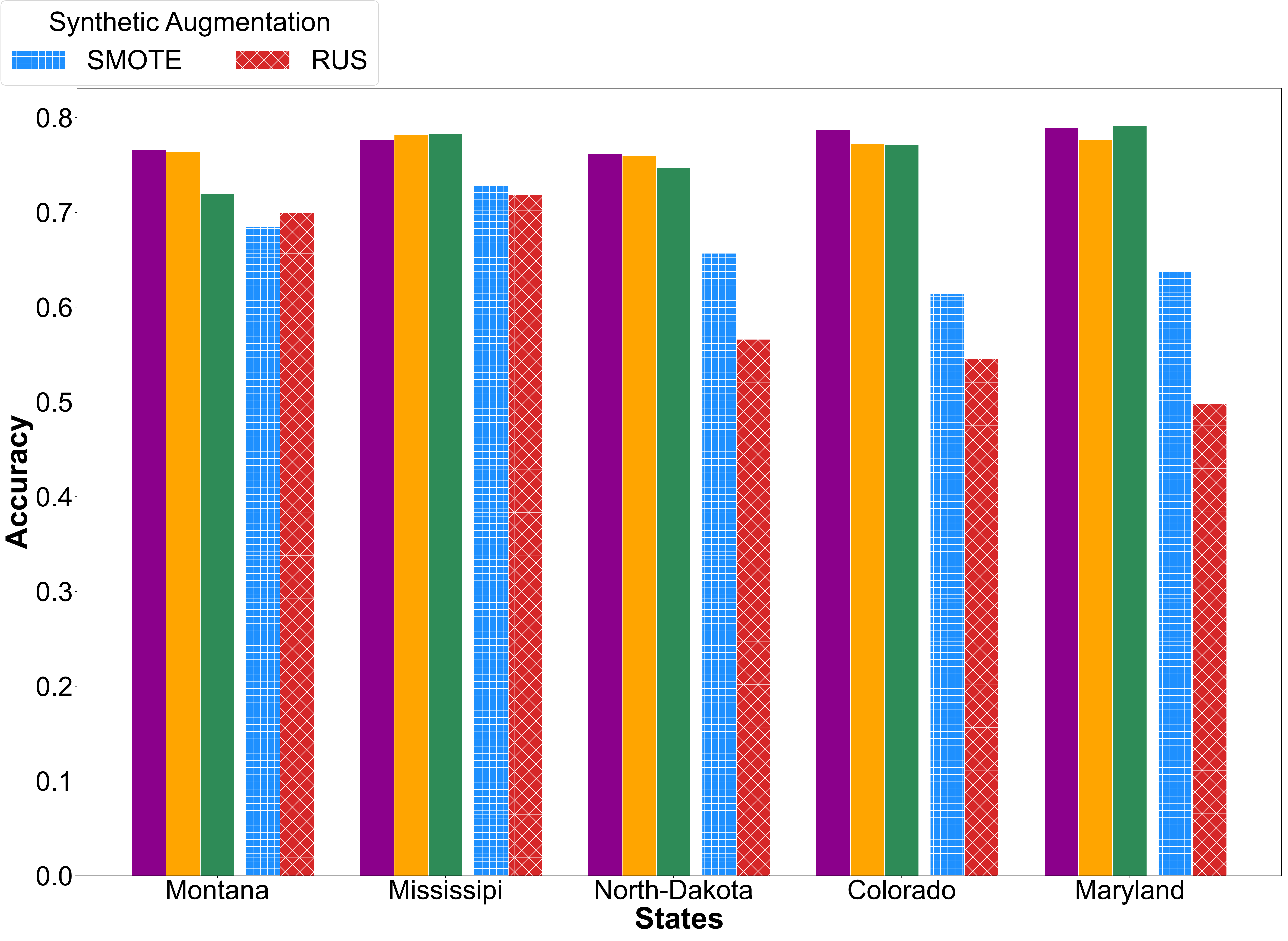}
       \caption{Accuracy of each method on each of the five states}
       \label{fig:acc}
    \end{subfigure}
    \caption{The results of a Logistic Regression model trained on each of the five augmentation methods and tested on the five states. 
    In \textbf{(a)} the Eq.Odds values of the five approaches are depicted. In \textbf{(b)} the Accuracy of the model w.r.t. each augmentation method for each state.}\label{fig:model_perf}
\end{figure}

In Figure~\ref{fig:model_perf} the performance results of a LR model trained on each of the augmentation methods and tested on the corresponding are illustrated. Note that, there are two legends where the first one represents the three augmentations (including our method MASC) that are based on real (genuine) data and the other represents synthetic augmentation methods. In Figure~\ref{fig:fair} the Eq.Odds values are shown where we can see the purple bar, representing our method MASC gets the best results for three states Montana, Mississippi, and North-Dakota as well as standing in the third best for two other states Colorado, and Maryland. Interesting observation comparing to results in Table~\ref{tab:augment} where SMOTE and RUS had better DI and SP results, it is observed that in model performance analysis, along with Geographical neighbors our method outperforms the SMOTE and RUS in all the states (Except for Mississippi where Geographical neighbors augmentations stands slighly worse than RUS) for both the metrics, accuracy and eq.odds. In Figure~\ref{fig:acc} where the accuracy results are compared, again the same situation is observed where MASC outperforms RUS and SMOTE and has the best accuracy in four of the states Montana, North-Dakota, Colorado, and Maryland and also stands in the third best for Mississippi.

\subsection{Discussion}
\label{sec:discussion}
From an analytical perspective although our method MASC seems to stand statistically comparable to or lower than the SMOTE and RUS in terms of DI and SP in Table~\ref{tab:augment}, but it outperforms both these methods in model performance results reported in Figure~\ref{fig:model_perf}. The reason for the former is because in our experiments, we implement a version of SMOTE and RUS that statistically augment protected groups, but still w.r.t. model performance measures, their augmentation is not comparable to real-world (genuine) data augmentations (e.g. MASC and Geographical neighbors). In that case, in Figure~\ref{fig:fair} and Figure~\ref{fig:acc} it is observed that MASC and Geographical neighbors (except for one case) outperform in all cases the two synthetic augmentations and once more we can highlight the importance/difference of real-world (genuine) data augmentation compared to synthetic/generated data. 

Moreover, there are ethical and technical issues with SMOTEing and RUSing for protected group imbalance augmentation. Starting with RUS: looking at Figure~\ref{fig:dist_original} only $0.36\%, and 1.35\%$ of the population belong to the minority group1 (Black) of the states Montana and North-Dakota, respectively. For the cleaned dataset it is no more than 20, and 60 instances each. So, with such a small number of instances, it is very unlikely for any learning algorithm to produce reliable predictions while being imposed to test data. This was also observed in Figure~\ref{fig:acc} where the Eq.Odds results of the RUS method always report one because it basically predicts all the under-sampled data to belong to the majority class. Subsequently, this lack of reliable performance might even get worse in cases where learning parameters are applied to out-of-distribution (OOD) data. An example of OOD is training on the augmented data (in our experiments are 2019 US-Census dataset) and then applying the model for future data, e.g. 2020, and later data of the same state. This is left as an open question to interested readers to test and analyze the results. Furthermore, another question is: what about inter-sectional groups when there is imbalance also w.r.t. more than one attribute; for example how would SMOTE and RUS perform if gender and ethnicity are studied simultaneously? For example, if only exists one instance of coloured-skin females within the 20 samples in Montana dataset, the algorithm will only learn to infer one class label among these group of instances which could lead to highly unreliable and deficient predictions on test data.

In case of SMOTE, it over-samples the minority group of 20 or 60 instances to generate hundreds of times more data. So, these synthetically generated data are only specifically applicable to this application because they need to be very carefully tailored for the application. This may describe the worse performance in Accuracy and Eq.Odds despite balancing the groups in training data perfectly (GR, DI, and SP measures) in the experiments. One of the limitations of SMOTEing is data types. How would it work with categorical data? One has to define a multi-valued vector of features and statistically over-sample the outnumbered categories while they are encoded numerically which results in severe performance deterioration because of much larger search space. However, our method is easily adaptable to categorical or other data types.

We would like to also highlight once again that in this study we only study the 2019 data so that conditions 2 and 3 of Section~\ref{sec:assumptions} do not apply to our analysis. In future works, it can be studied also where the distribution of target class (condition 2) or when the decision boundary changes (condition 3) which can happen when analyzing different historical records for each state, e.g. comparing the 2014-2019 data of each state. Also it is worth mentioning that there is a lack of similar datasets especially from the European countries that can be provided for research which can open up space for more studies in this direction.

\section{Conclusion}
\label{sec:conclusion}

In this paper, we propose a spectral clustering-based methodology to tackle data representation and protected-attribute group-imbalance biases. The motivation for developing this pipeline is to utilize contextually similar but separate datasets coming from similar sources, to augment one another in order to provide unbiased or less biased training sets using shared instances from contextually similar neighboring datasets. Our \textit{MASC} approach identifies an optimal number of clusters based on inherent similarities of the input tasks and clusters them according to a robust and scalable MMD two-sampled test. Furthermore, it categorizes similar tasks based on their pairwise distribution discrepancies in a kernel-based affinity space. Experimental results on \textit{New Adult} datasets reveal the promising performance of the proposed \textit{MASC} in dataset debiasing and superior performance in improving predictive and fairness of learning models trained using the augmented training sets obtained by it. Moreover, it is preferable over synthetic data augmentation methods such as SMOTE and RUS since it augments based on genuine (real) existing data in contrary to the synthetic ones which are usually used under many ethical concerns. In future work, we will study the effect of normalized spectral clustering on the size and shape of clusters produced. We also encourage to extend our analysis to temporal aspects of the datasets by assuming change in the conditional probability of outcomes \emph{$P(Y|X)$} in \emph{$X\rightarrow Y$} problems for each year of the input datasets. Another interesting study would be to compare our method and the Geo-neib with a version of the SMOTE and RUS for multi-class imbalance or regression problems where the targets are multi-class or continuous.

\begin{acknowledgments}
This work has received funding from the European Union’s Horizon 2020 research and innovation programme under Marie Sklodowska-Curie Actions (grant agreement number 860630) for the project ‘’NoBIAS - Artificial Intelligence without Bias’’. This work reflects only the authors’ views and the European Research Executive Agency (REA) is not responsible for any use that may be made of the information it contains.
\end{acknowledgments}


\bibliography{Bibliography}

\end{document}